\documentclass[smallextended]{svjour3}
\usepackage{graphicx}
\usepackage{listings}
\journalname{Autonomous Agents and Multi-Agent Systems}

\title{Introducing the Talk Markup Language (TalkML)}
\subtitle{Adding a little Social Intelligence to Industrial Speech Interfaces}

\author{Peter Wallis}

\institute{Centre for Policy Modelling\\
	Manchester Metropolitan University Business School\\
	All Saints Campus, Oxford Road\\
	Manchester M15 6BH, UK\\
}

\date{}

\begin{document}

\maketitle

\begin{abstract}
Virtual Personal Assistants like Siri have great potential but such
developments hit the fundamental problem of how to make computational devices
that {\em understand} human speech.  Natural language understanding is one of
the more disappointing failures of AI research and it seems there is something
we computer scientists don't get about the nature of language.  Of course
philosophers and linguists think quite differently about language and this
paper describes how we have taken ideas from other disciplines and implemented
them.  The background to the work is to take seriously the notion of language
as action~\cite{austin,searle69} and look at what people actually {\em do} with
language using the techniques of Conversation
Analysis~\cite{sacks,HutWoo98,tHave99}.  The observation has been that human
communication is (behind the scenes) about the management of social relations
as well as the (foregrounded) passing of information.  To claim this is one
thing but to implement it requires a mechanism.  The mechanism described here
is based on the notion of language being intentional -- we think intentionally,
talk about them and recognize them in others -- and cooperative in that we are
{\em compelled} to help out~\cite{tomasello08}.
The way we are compelled points to a solution to the ever present problem of
keeping the human on topic.  The approach has led to a recent success in which
we significantly improve user satisfaction {\em independent} of task
completion~\cite{wcl14} and the work described here shows how the approach
might impact on commercial interests.  VoiceXML is a default industrial
standard used to script automated telephone conversations and is based on the
notion of using forms to update information state.  Talk Markup Language
(TalkML)
is a draft alternative to VoiceXML that, we propose, greatly simplifies the
scripting of interaction. By embracing the intentional nature of language and
attending to its social side, we provide default behaviour for the two VoiceXML
grammar exceptions, {\tt nomatch} and {\tt noinput}.
\end{abstract}


\section{Introduction}
Scripting dialog to make a movie is one thing but scripting interactive dialog
for use ``in the wild'' is another.  Siri~\cite{siri}, Amazon
Echo~\cite{aecho}, JiBo~\cite{jibo} and Nina~\cite{nina} might be promoted by
using movies that demonstrate the concept but spontaneous human interaction
with such systems is another thing and is unlikely to result in happy users.
There is an interesting problem here that has historically not suffered from a
lack of interest or resources.
From the 1950's there has been considerable financial backing for machine
translation and the infamous
Microsoft desk-top assistant (the Paper-Clip) had both the resources and the
user base to develop whatever could be imagined. Siri again has the resources
and this time there is certainly better ``image management'' but people still
get frustrated with spoken language interfaces and how to fix this is an
unsolved issue.
The assertion here is that although we humans think we might be only interested
in the information, behind the scenes we cannot help but attend to a whole
range of cues that machines just don't get right.

There is of course on-going interest in looking at dialog systems, ranging from
languages for scripting dialog such as AIML~\cite{alice} for chatbots such as
Ikea's Ask Anna~\cite{ikeaECA} and Voice XML~\cite{vxml} for interactive
voice response (IVR) systems, through conventional plan-based
systems~\cite{trains}, BDI architectures for
dialog~\cite{ArdBoe98,wmod01,boella02,LeKrWa04,wobcke07}, add-hoc state-based
systems~\cite{wilksEtAl2011} and machine learning systems based on partially
observable markov decision processes~\cite{pomdp07,hwpomdp}.  The work
described here extends that done using BDI, the Belief Desire and Intention
agent architecture~\cite{BIP,RaGe95,wool00}, for dialog management as, we claim, the key to a solution is to address the intentional nature of language.

Talk of intentions is contentious as many consider such talk to be imprecise 
and too much ``in the head''.  Indeed, as discussed below, intentions
are explicitly rejected by those working with Conversation Analysis (CA) which
has beem problematic as CA is also a central thrust of the work described here.
It is true that intentions are {\em ascribed} to things by some to explain
behaviour and such ascription is highly unscientific:  electricity {\em wants}
to take the
shortest path but ``gets confused,'' or thinking that a chess playing computer
(using a search algorithm) {\em intends} to take my rook~\cite{Denn87}.  One
may know how the computer chooses its next move, but that does not stop one
reasoning about its behaviour in terms of intentions.
On the other hand most of us are fairly sure that my son intending to push his
sister is different to him accidentally pushing her because there really is an
intention there somewhere.  What is more, when we program computers to act
intentionally, we can actually put data structures in there that correspond to
intentions. This type 2 intention is objectively real -- as real as software
gets -- in the case of the such computer programs, and is also real (let us
assume) when we talk about my son wanting an ice cream.
Finally the intentional model of others is built in to language.  Even those
who do not want to talk of intentions have to use language that assumes people
have type 2 intentions.  If someone describes the reason for an action as
being ``to show the result of a search''~\cite{Crab03} we are {\em already}
talking about an intent.  Indeed intentional structures in text have been
discussed before, most notably Grosz and Sidner~\cite{GrSi86} but intentions are
also first order objects in that they are talked about.  Consider ``I'll just
put that in the system for you; is that alright?''~\cite{wmod01}.   The
intention in ``Would you like to go to the pictures tonight?'' is just as much
``on the surface'' as any factual content in something like ``Fellini's
$8\frac{1}{2}$ is showing at the Showcase Cinema tonight''.  For CA, it has
been very productive to ban talk of ``things in the head'' but when throwing
out type 1 intentions they seem to have also thrown out type 3 intentions which
would seem to be as much ``in the text'' as anything else.

The author agrees with the CA community that the wide spread practice of
assigning ``dialog acts'' such as {\sc query-yes-no}~\cite{carCL97,swbd-damsl}
to text segments~\cite{BaGo97} is problematic~\cite[p289]{pragmatics}. Such
tagging schemes slip a scientist's theory in to the analysis with very little
justification~\cite{hovy10}.  To say ``Fellini's $8\frac{1}{2}$ is showing at
the Showcase tonight and I thought I'd go.'' is as much a yes/no question as
the above in the right context.  In fact it would need to be categorised as a
yes/something-else question as an unadorned response of ``no'' would say far
more than that she didn't want to see Fellini tonight.  Of course these
subtleties usually do not occur in conversation about arranging flights or
booking a taxi but when things go wrong, humans have the tools for dealing with
such subtleties to hand and cannot help but use them.

The work described here looks specifically at interactive voice response (IVR)
systems but can be (and is being) applied to other domains such as virtual
personal assistants (VPA) in the spirit of Jibo and Amazon's Echo.  Although
VPAs are certainly more cool than telephone banking or calling for a taxi, IVR
systems are deployed now with millions of users and considerable 
commercial interest in the task of customer service provision.  Talking to a
machine over a phone also mitigates a range of problems for the speech
recognition including issues with far-field microphones and with notions of
attention~\cite{wallis10}.  Note also that the point -- improved authoring by
looking again at the role of {\tt nomatch} and {\tt noinput} exceptions in
VoiceXML -- should be seen as {\em an instance} of the more general solution of
using the tools of Conversation Analysis to look at interaction.

\section {Dialog as action in a social space} \label{daass}
The conventional wisdom is that natural languages are the
definitive symbol system and computers are, in a very literal sense, {\em
universal} symbol manipulation systems. So what could we be possibly missing
when it comes to conversational machines?
The answer has been of course the notion of agency and situated action.
Computers can do something other than manipulate symbols; they can implement
arbitrary causal relations between sensors and actuators.  Computers can also
implement thermostats, and behaviour based robotics has certainly made some
significant progress over good old fashioned AI systems (GOFAI) that sense,
model, plan and act~\cite{brooks91}.
Applied to language understanding -- and in particular dialog -- the success of
situated action suggest we take seriously Austin's notion of language as
action~\cite{austin,searle69}.  This has led the author to look closely at the
linguistic camp known as Conversation
Analysis~\cite{sacks,Firth96,PomFehr97,HutWoo98,tHave99,seedhouse04,BnS07} in which the focus
is on the ``work done'' by an speaker in making an utterance. Rather than
looking in heads for meaning, we need to look at the relationship between the
head and the world around it.

\begin{figure*}[t]
  \includegraphics[width=0.50\textwidth]{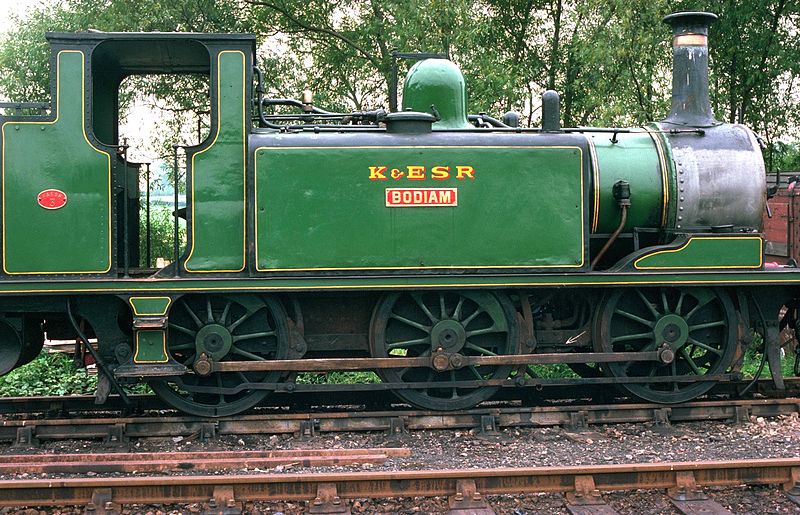}
  \includegraphics[width=0.50\textwidth]{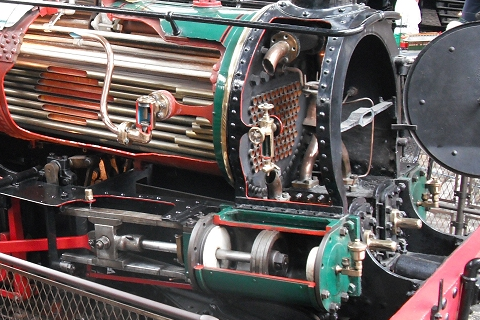}
  \caption{Understanding how steam engines work is best done by looking
  inside.} \label{steamEngine}
  \includegraphics[width=0.44\textwidth]{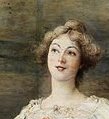}
  \includegraphics[width=0.56\textwidth]{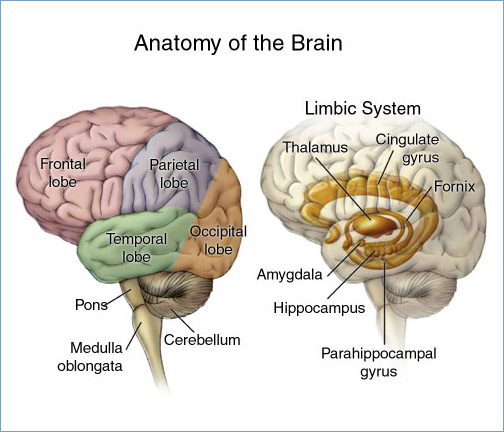}
  \caption{Understanding how heads work is often tackled the same way.}
  \label{brains}
\end{figure*}
Consider the classic reductionist approach to understanding how something
works. Figure~\ref{steamEngine} (b) certainly helps us understand how a steam
engine works, and ``looking inside'' is certainly a popular approach when it
comes to cognitive science (Figure~\ref{brains}).
\begin{figure*}
  \includegraphics[width=0.58\textwidth]{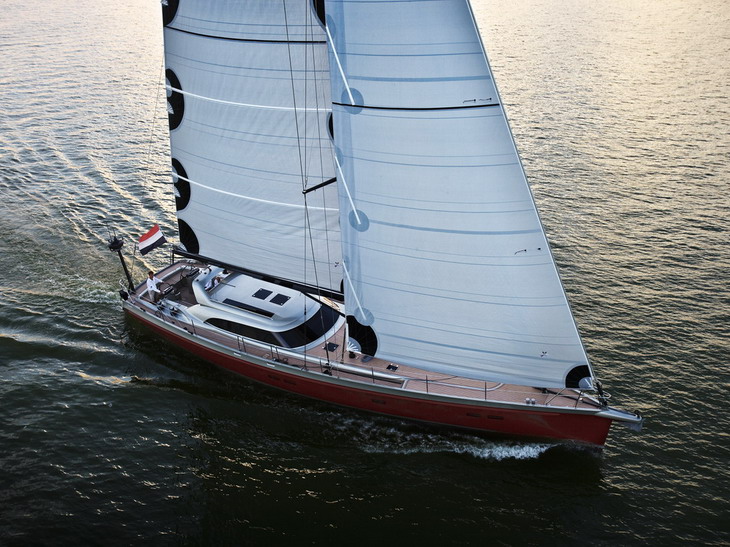}
  \includegraphics[width=0.42\textwidth]{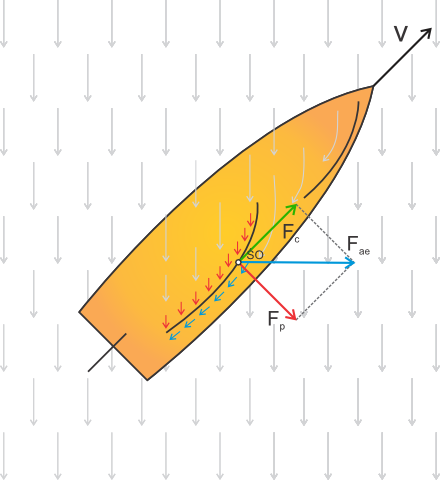}
  \caption{Understanding how sailing ships work is best done by looking
  outside.}
  \label{sailing}
  \includegraphics[width=0.40\textwidth]{800px-Francesco_vinea_head.jpg}
  \includegraphics[width=0.60\textwidth]{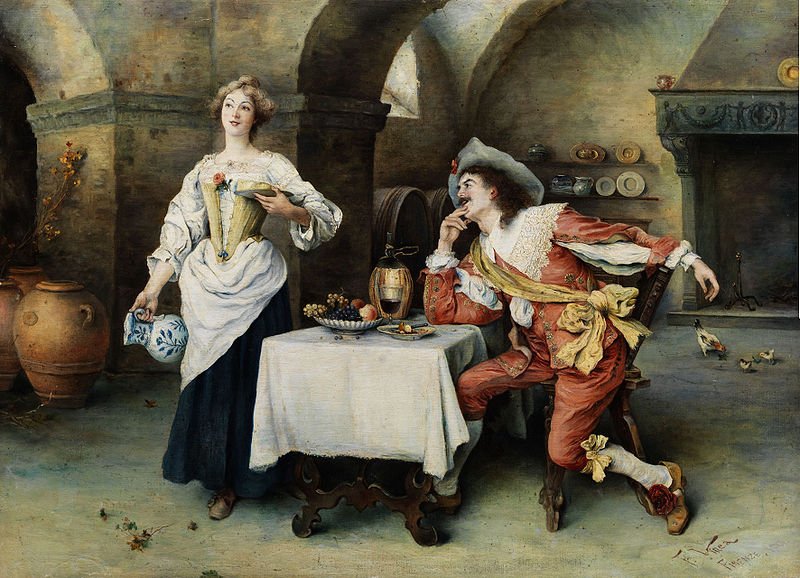}
  \caption{Understanding how heads work can be tackled the same way.}
  \label{flirting}
\end{figure*}

Taking the same approach to understanding sailing ships is however
not useful -- indeed the inside of a racing yacht is virtually empty. To
understand sailing, we need to look at the relationship between the object of
interest and its environment(Figure~\ref{sailing}). The ``work done'' by the
sails is to transfer the force of the wind to the vessel, while the keel
ensures that only a component of that force is translated into motion.
When it comes to heads, in order to understand how people work, a valid reductionist approach is to take a step
back and look inside the system of which a head is part.  The notion of 
semantics and the body (~\cite{ruthrof97} among many) has been an interesting
approach but Figure~\ref{flirting} (b) is a head attached to a body embedded
in a web of social relationships, not all of which are in the scene.  As
``members of that community of practice,'' we understand the communication
going on here. We understand -- litterally -- the ``work done'' by his
posture, their gaze, and of course the speech acts they produce.

One might expect that in language the work done is done by words and that a
big book of words listing the work done by each would be a reasonable approach.  However
consider this classic example from the CA literature in a doctor/patient
conversation:

        {\bf Patient:} {\it this treatment, it won't have any effect on us
        having kids will it?}

        {\bf Doctor:} [3.2 seconds silence]

        {\bf Patient:} {\it It will?}

        {\bf Doctor:} I'm afraid so..

\noindent 
Although it might seem reasonable to consider words to have meaning that can be
looked up independent of context, the same is not true of silence and
in the example that is certainly a meaningful silence. However meaning is
determined here, the mechanism is hard at work {\em also} when we figure out the meaning of words. 

Looking through the CA literature, it would seem human communication is is full
of normative behaviours -- rules that can be broken, but to break them will be
interpreted, not only by one's conversational partner, but also by any
audience~\cite{laurel93}.  These rules are behind the scenes in that we do not
consciously think about them; we know they are there and think about dealing
with them but their existence goes largely unnoticed.
They are also ubiquitous.  Making an apology is a complex
process~\cite{owen83}, but then so is saying goodbye~\cite{schsac73}. To say to
a friend while standing at a bar ``can I borrow 50p?'' is acceptable; to ask
without preamble or explanation ``can I borrow 500 pounds?'' is not.
Why~\footnote{If there are readers who think it is obvious and needs no
explanation, please note there are cultures where such a question is perfectly
polite, and one appropriate polite response is an unadorned or elaborated
``No.''}?

Note this is not scientific knowledge but rather ``folk'' knowledge.  As with
other ethnomethods~\cite{garf67}, the point is to capture the every-day
knowledge that people use to do what they do.  To do this effectively CA has
disavowed conjecture about the mechanism or ``rules in the head'' that might
have general applicability, but rather focus on what happens in particular
instances of communication and what observable behaviour contributes to choices
made.  The idea that language use requires folk knowledge may be obvious but
the extent to which folk knowledge is core is perhaps borne out by the success
amateurs have in developing conversational agents for things like the Loebner
Prize~\cite{loebner}.  Folk know exactly, in context, what to say.
What the untrained do not know is how to abstract from the surface form of an
actual apology say, to something that can generalize across different
contexts.  Indeed 50 years of NLP research suggests experts do not know how to
do that abstraction either.

\section{How language works}

CA is a methodology that enables researchers to notice what is going on and the
approach has been very effective at collecting samples of language in action
and seeing how they work.  It is strong on methodology and shy on theory but
Seedhouse~\cite{seedhouse04} gives a summary of ``the findings of CA over the
last 50 years''.  The summary is that a speaker's utterance will go either:
\begin{description}
\item[Seen but unnoticed] \hfill \\
if it is the answer to a question or a greeting in response to a greeting and
so on.  If the speaker produces the second part of an adjacency pair, then the
conversational partner (who produced the first part of the pair) will not
notice the utterance and will move on.  If the utterance is not as expected
it might be

\item[Noticed and accounted for] \hfill \\
If the speaker says an utterance that is not expected - not the normal
response -- the addressee does not instantly give up, but actually works hard
to figure out why he or she said what was said.
As a classic example consider some one walking in to a corner shop:

\begin{quote}
\begin{tabular}{r|p{60mm}}
\textbf{A:} & Hello.  Do you sell stamps?\\
\textbf{B:} & First class or second class?\\
\end{tabular}
\end{quote}

\noindent

Unless it is pointed out, people often do not notice that B's response is not
an answer to the question.  B's response can however be accounted for  -- a
process we humans are willing to work hard at~\cite{EggSla97}. If the conversational partner cannot account for the utterance then the speaker

\item[Risks sanction]\hfill \\
It seems humans have a notion of fairness and feel justified in sanctioning
the speaker if they think the speaker is not cooperating in the communication
process.  This is not a prescriptive rule taught to well bought up children; it
is descriptive of what people do.  The form of the sanction depends on many
things and may be highly culturally dependent, but this is where notions of
power and distance, roles and expectations comes into play.
\end{description}
This process of working through the seen-but-unnoticed,
noticed-and-accounted-for, to sanction, is not something we think about; it is
just what we do and as such it is hard to notice in action.  What is more,
humans rarely get to the third stage of sanctioning.

Conversational agents are another matter.  Consider the
conversation in Figure~\ref{smb} between a human and a chat bot.
\begin{figure*}
\begin{tabular}{rl}
  PROGRAM:& G'day mate.\\
  JUDGE02:& G'day. How's it going?\\
  PROGRAM:& My mother once said "How is it going is a mystery." She
           was a bit nuts.\\
  JUDGE02:& Uh-huh.\\
  PROGRAM:& Are you still learning to speak, my friend?\\
  JUDGE02:& NO, no. But your anecdote took me a bit by surprise.\\
  JUDGE02:& How was your morning?\\
  PROGRAM:& Oh well, that's a pity!\hspace{10mm} $\leftarrow (1)$ \\
  JUDGE02:& I'm sorry; what's a pity?\\
  PROGRAM:& Apologies accepted.\\
  JUDGE02:& I feel compelled to insult you, and we've only been
           talking a minute.\\
  PROGRAM:& We are bloody well talking, aren't we?\\
  JUDGE02:& Oh baby. We sure are. So here are some keywords for
           you: shit mother baseball.\\
  & ...\\
  \end{tabular}
  \caption{A conversation with the winning entry for the 1997 Loebner Prize
  competition. Most were much better than this.} \label{smb}
\end{figure*}
On a first pass what stands out is the way the conversation simply falls
apart.  One can point to the place where things start to go wrong, but for
a considerable number of turns, the human at least is working toward repairing
the interaction with apologies and warnings.  At $(1)$ the machine utters
something that the judge cannot account-for. The judge tries to get the
machine to explain, which fails, resulting in an explicit threat of sanction.
In the end the threatened ``keywords'' include swearing even though the judge
is well aware that the world is watching.  A standard response to this example
is to think we just need to ensure the machine does not say anything that
cannot be accounted-for but note that 4 lines prior to $(1)$ the human says
something the machine (acts as if it) cannot account-for and the human's
response is quite different.  The events are mirror images of each other but
the human's handling of the situation is so automatic for us that it is
often hard to notice the symmetry. Human language use is situated action in an
environment.  Getting machines to do conversation, the context is just as much
part of the process as the code and that context is full of highly socialized
people.

This example illustrates clearly the process a human goes through. Less
clear is perhaps the notion that the reader, as a member of the community of
practice~\cite{tHave99} has access to information about the work being done by
utterances in the transcript.  For instance the reader is likely to think the
judge is ``within her rights'' to sanction as she does.
Indeed, if she didn't sanction such behaviour, we might think that she also is
a machine.
Although expert knowledge based on introspection is questionable, one's own
every day knowledge of the ``meaning'' of utterances is the raw material for
the CA methodology. This notion, that she is ``within her rights'' is an
objective fact that the (trained) scientist can observe about his or her
unscientific self.  One might be tempted to decompose such observations into
more abstract notions such as Face Threatening Acts~\cite{BnL}, but that
visoral response, be it an emotional response or knowledge, is the ``work
done'' by the words as spoken.
The key is that that visoral response is shared by the ``community of
practice'' of language users.  One might do experiments on other members of the
community (as is done is Section~\ref{swoc} below) but when the researcher is
also a member of that community, his or her response to an utterance is valid
objective data, as long as the researcher's specialist knowledge does not get
in the way.
Of course one's response may not be exactly the same as another's but it will
be close most of the time -- otherwise communication would not work.  When it
is not close, the expression risks sanction and initiates a repair process.
It is this process of working through the
seen-but-unnoticed/noticed-and-accounted-for/risks-sanction that machines do
not get right and need to if they are going to participate in communication
with humans as humans do it.

\subsection{Keeping humans on topic}
There is no doubt human heads contain wide and general knowledge compared to
your average computer program. In the early days -- as humans always do - we
projected our desires onto the unknown and computers of the future were
expected to be infalable and omnicient~\cite{forbiddenPlanet,deskSet}.
In practice machines have limited knowledge and deliniating the boundaries of
that knowledge is problematic.  If however we embrace the idea that human
communication involves (the threat of) sanction, a machine participating in
human communication can be {\em expected} to sanction in much the same way as
we expect the judge to sanction in the example in Figure~\ref{smb}.
The proposal is that machines should sanction when such action is deemed {\em
reasonable} by the community of practice.  This provides a means of keeping the
user on topic and behaving appropriately so that speech recognition can work
and the knowledge the machine has can be useful.

The acceptance of sanction assumes the user is participating in a ``willing
suspension of disbelief'' and pretending to talk with a person\footnote{A pretense that comes naturally to humans using computers~\cite{ReeNas96}.}, who is in fact  machine --- a machine pretending to be a person.
This pretense is not the only way to make speech interfaces and
indeed a popular approach is to to treat speech as simply another modality for
classic HCI practice~\cite{ibtbagm}.  The HCI approach has been to have the
user see the machine as a tool~\cite{HCIintent} and to make sure the user knows
how to wield the tool and making the consequences of action
obvious~\cite{srp07}. This approach certainly has the advantage that
blaming the tools for our mistakes is unjustified, and we user's are compelled
to feel we {\em ought} to find out how to use the tool better by, perhaps,
reading the manual.  Some times the tool is however actually broken.  Language
as used by us social actors is a means of negotiating blame rather than just
blaming the user. If the machine can participate in this negotiation, the
machine can negotiate the limits of its conversation.  Using CA, we can look at
how humans negotiate what is reasonable and hopefully make a machine behave the
same way.

In 2001 during some wizard of oz trials the author tried a breaching
experiment~\cite{garf56} and phoned the system to book a table for two for
that evening.  The system however was intended to provide directions about the
University campus.  The wizard's response was to say {\em ``You have called the
University of Melbourne; How can I help?''} This ``move'' on behalf of the
wizard suggests that the request is inappropriate for her role.  Compare the
work done by what she said with ``I'm Tanya; How can I help?''
In another case an automated taxi booking system holds a conversation in which
the dialog is slightly tense and in the background the caller can hear a busy
call centre.  To the caller the ``person'' on the other end of the phone line
sounds very busy and efficient and one is certainly discouraged from making rambling conversation about the weather, why you want to get a taxi or to try experiments with the system's speech recognition capabilities.
In both cases the role of ``the system'' delineates what it is
``reasonable'' to expect and of course if the human is not reasonable, the
system could sanction {\em and be viewed as justified in doing so}.  This
feature of language use is exploited below, but first we need to look at {\em
how} a machine might be made to behave this way.

\section{Inside (synthetic) heads}
Conversation Analysis tells us what a conversational agent needs to do, and
what it needs to do is far more complex than simply providing the facts, but
is not helpful when it comes to deciding {\em how} the required behaviour is
implemented.  Indeed the CA community appears to explicitly rule out any such
discussion and it is informative to note how Seedhouse words his summary as a
set of ``findings'' rather than as a set of (descriptive) rules that might be
construed as (prescriptive) rules in heads.
This is perhaps a consequence of being developed at around the same
time as psychology was toying with behaviourism but whatever the reason, 
their science has made a feature out of {\em not} admitting explanations involving mental attitudes - a feature which has been highly productive.
With an unwavering eye on the Engineering however, if we are going to
implement an artifact that can participate in the process of conversation, we
need to put {\em something} inside a synthetic head.  From an Engineer's
perspective we are interested in what is inside even if they are no more
complex than the wiring of a thermostat or the rudder chains of a sailing ship.

\subsection{Intentional (explanations of) behaviour}

For the seen-but-unnoticed, the classic shallow constructs used in
chat-bots and VXML work well --- specify what is expected as an answer with a
regular expression or a speech grammar, say a question, then update the
information state from the answer template via some (possibly minimal) semantic
interpretation.  The (threat of) sanction also seems to fairly straight forward to implement.
The difficult part is  accounting-for a user's utterance when it is not the
expected response and this would appear to require full human knowledge.
What is more, and in contrast to CA's methodology if not it's theory, it
requires thinking about type 2 intentions.

Consider this (naturally occurring) example from Eggins and Slade talking
about sequential relevance:
\begin{quote}
\begin{tabular}{r|p{60mm}}
\textbf{A:} & What's that floating in the wine?\\
\textbf{B:} & There aren't any other solutions.\\
\end{tabular}

\noindent
You will try very hard to find a way of interpreting B's turn as somehow an answer to A's question, even though there is no obvious link between them, apart from their appearance in sequence. Perhaps you will have decided that B took a common solution to a resistant wine cork and poked it through into the bottle, and it was floating in the wine. Whatever explanation you came up with, it is unlikely that you looked at the example and simply said `it doesn't make sense', so strong is the implication that adjacent turns relate to each other~\cite{EggSla97}.
\end{quote}
This example demonstrates just how hard human communicators are willing to work
at {\em recognizing intent} in the speech acts of others.  By contrast consider
this example from Mann~\cite{Mann88}:
\begin{quote}
\begin{tabular}{crl}
  1& Child:& I'm hungry \\
  2& Mother:& Did you do a good job of your geography homework? \\
  3& Child:& Yeah. \\
  4& Child:& What's to eat? \\
  5& Mother:& Let me read it. \\
  6& Mother:& What is the capital of Brazil? \\
  7& Child:& Rio de Janeiro. \\
  8& Mother:& Think about it. \\
  9& Child:& It's Brasilia. \\
  10& Child:& Can I eat now? \\
  11& Mother:& I'll let you have something later. \\
  12& Mother:& What is the capital of Venezuela? \\
  13& Child:& Caracas. \\
  14& Mother:& Fine. \\
  15& Child:& So what can I eat? \\
  16& Mother:& You want some cereal? \\
  17& Child:& Sure. \\
\end{tabular}
\end{quote}
Mann uses the example to indroduce the idea of a dialog game but it also
demonstrates just how easy we humans find it to recognize the intent
of others.
%Intention recognition cannot be treated as a plug-in; it is an essential feature of communication as humans do it.
In this conversation it is not until line 11 that Mother uses any words or
concepts that might be semantically linked with the Child's opening statement,
and indeed the words she speaks are from an entirely different context.  Of
course the child knows what mum {\em wants} and it is only because the child
knows what mother wants that her utterances make any sense as part of this
conversation.
If one is not convinced that mother's desires are a factor, consider a
conversation in which at line 2 Mother says ``I rode my new bike thirty five
kilometres today''.  Being a member of the community of practice one can put
ones self in the shoes of the child (breeching experiments being unethical
today) and imagine one's confusion. One may indeed ``try very hard to find a
way of interpreting'' riding a bike as relevant to feeding a child -- indeed it
is expected -- but the point is that one's explanation will involve unravelling
Mother's reasoning about bikes and feeding children. That is, one will be
trying to understand what Mother {\em intended} by her utterance about bikes.
These two examples demonstrate Tomasello's claim~\cite{tomasello08} that human
communication is {\em intentional} and {\em cooperative}.  We humans read off
the intent of others -- it may be difficult to recognize the intent of Hitler
invading Poland, but seeing two children tugging at a teddy bear the human
observer will be quite sure they both {\em want} it~\cite{Denn87} -- and we are
willing to work hard at it.  The great apes, although also good at recognizing
intent would look at the Eggins and Slade example, decide it didn't make sense
and move on.

\subsection{Plan based dialog systems}
One reason for embracing the notion of intent is that there is an existing
model of reasoning with intentions that provides a cheap and simple means of
doing (limited) intention recognition.  The empirical question is whether this
limited form is effective.

In the early days of AI it was assumed intelligent machines would think the way
(we think) we think. Much of what has been referred to as ``good old fashioned
AI,'' or GOFAI, was based on the idea of formulating a {\em plan} to achieve
{\em goals} and that was seen as the only thing for an intelligent being to do.
To walk, I would (subconsciously) plan to lift my leg and move my foot in the
direction I wanted to travel.  These systems could be quite inflexible and
slow and have given way in robotics to layered architectures
in which reactive behaviours interfere with each other~\cite{brooks91} or are
managed by higher level layers~\cite{Ark98}.
Criticism of the plan based approach to dialog~\cite{wilksEtAl2011}
can often confound the limitations of GOFAI planning with the use of plans.

The observation is that the popular chat-bot mechanism first described by
Weizenbaum~\cite{eliza} with his ElIZA program has a strong resemblance to the
reactive layers in modern robotics.  This approach gives the system the ability
to respond to changes in the environment but does not, by itself, provide a
means of committing to long term action.  Long term action is often addressed
by maintaining some idea of ``state'' (for example \cite{lbckw97,creer11}) but
changing state is problematic and usually left to the research assistant to
figure out.

Another approach to the problem of situated action was the introduction of BDI
architectures based on the management of plans rather than
planning~\cite{BIP,RaGe95,wool00}. This approach embraces the idea of
reasoning in terms of commitment to plans (there are other ways to reason)
but in practice does it by working from a library of pre-defined plans.
Figure~\ref{acl} from Wooldridge~\cite{wool00} outlines an algorithm that
balances commitment to plans against responding to a changing environment.

\begin{figure*}
Algorithm: Agent Control Loop Version 7
\begin{lstlisting}
1.
2.  $B := B_0$   /* $B_0$ are initial beliefs */ 
3.  $I := I_0$   /* $I_0$ are initial intentions */
4.  while true do
5.    get next percept $p$;
6.    $B := brf(B,p)$;
7.    $D := options(B,I)$;
8.    $I := filter(B,D,I)$;
9.    $\pi := plan(B,I)$;
10.   while not $(empty(\pi)$ or $succeeded(I,B)$ or $impossible(I,B))$ do
11.     $\alpha := hd(\pi)$;
12.     $execute(\alpha)$;
13.     $\pi := tail(\pi)$;
14.     get next percept $p$;
15.     $B := brf(B,p)$;
16.     if $reconsider(I,B)$ then
17.       $D := options(B,I)$;
18.       $I := filter(B,D,I)$;
19.     end-if
20.     if not $sound(\pi,I,B)$ then
21.       $\pi := plan(B,I)$;
22.     end-if
23.   end-while
24. end-while
\end{lstlisting}
\caption{An agent that attempts to strike a balance between boldness and
caution. (From Wooldridge 2000~\cite{wool00})} \label{acl}
\end{figure*}
Taking this approach one does not have to have planning ``all the way down''
and indeed plans in the library may contain sets of ELIZA-like pattern-action
rules that simply ``produce behaviour'' that an agent might have if it had the
relevant goal.
Using this mechanism for dialog clearly distinguishes between normative
behaviour in which people just do things like respond to greetings with
a greeting and answer questions, and planned language use in which things are
said with an identifiable goal in mind.   The primary advantage of taking this
approach is that the explicit management of commitment to plans allows a BDI
dialog system to provide {\em mixed initiative at the level of
intentions}~\cite{intentmap}.  TalKML takes the BDI model of managing
commitment and, rather than ELIZA-like pattern-action rules, uses
VoiceXML grammars and prompt statements to provide the semantics/action. 

\subsection{User Satisfaction {\em without} Task Completion} \label{swoc}
The claim is that human language use is intentional and cooperative which is
more than just doing what one is asked to do.  To the utterance ``can you pass
the salt?'' people do not respond ``yes'' but rather account-for the question
by assuming the speaker {\em wants} the salt.
The general assumption has been that users don't like talking to a machine when
they phone the bank; don't like using desk top assistants and get frustrated
using embodied conversational agents on websites, because these systems do not
provide the right information.
But not being able to provide the right information is often inevitable --
human service providers regularly fail but do not, usually, get the same level
of abuse.  Rather than expecting computers to be perfect and working toward
that, our aim was to make a machine fail in the same way as humans do.

So how do human service providers negotiate failure?
The answer is that many talk about how they have tried to help.
This lead us to look at service provision systems, not as a tool
for accessing information~\cite{ibtbagm}, but as a social actor that {\em
wants} to help.  Rather than providing a tool that does exactly what is asked
of it in a timely manner~\cite{srp07}, we created a system that talked about
its plans to help the user achieve his or her goals.

We set up an experiment to compare user satisfaction with two telephone based service provision systems, both of which
provided the same information, but one of which talked about its plans. 
In order to keep the information provided the same, all the
techniques it tried failed, but it talked about it anyway. The task was to get
the phone number for a member of staff in the Department, and task completion
was pegged at 20\% - four out of five scenarios meant the subject was going to
fail to get the person's number using either system.  A phone answering
system was set up using an evaluation copy of the Small Business Bundle from
Nuance~\cite{nuance} and we recruited a number of subjects from the
institution's student population.
Our aim was to run a preliminary trial in order to make
sure we had everything in place for the planned experiments before purchasing
the system from Nuance.  It is true no sensible person would design an
experiment with such a small sample, but the aim was not to produce a
statistically significant result at this stage.
User satisfaction was measured using the User Experience
Questionaire~\cite{ueq} which had been designed for exactly this kind of
comparison although it is designed for new graphical user interfaces rather
than speech interfaces.  The questionaire is short, designed to capture the
immediate experience of the user, but was developed by running factor analysis
over a large set of questions from a range of HCI experts and over established and new interfaces.
\begin{figure*}
\includegraphics[width=\linewidth]{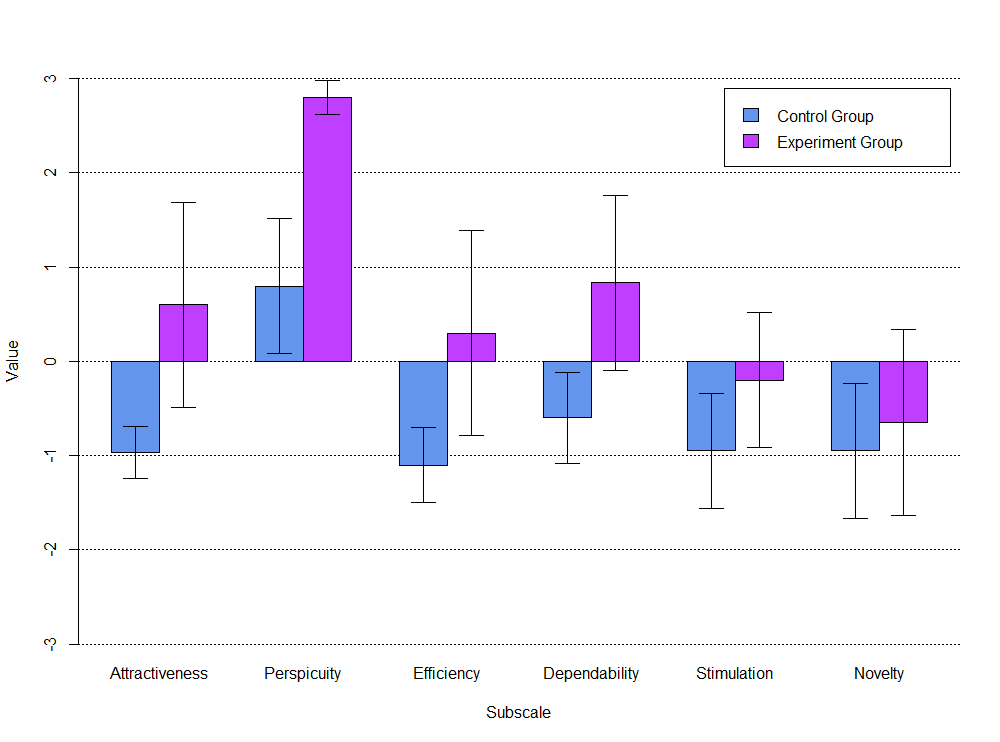}
\caption{A comparison using the User Experience Questionaire~\cite{ueq} to
compare two service provision systems, one of which is seen to try.}
\label{barplot}
\end{figure*}
Figure~\ref{barplot} compares the two systems on the six dimensions from the
UEQ process.  The labels on the dimensions are fairly self explanatory and the reader is directed to the original paper for the details.
The error bars indicate we have a statistically significant result (0.95)
and this means there is a one in twenty chance that we would not get the same
result if we re-ran the trial on a new set of subjects from the population
represented by the sample.  Given the statistics we do not intend to run this
particular trial again but instead are moving on to trials with targeting
a more generic population and with a task more in line with commercial
interests.

\section{XML for dialog}

Practical dialog systems tend to be based on the notion of information state
update~\cite{isu} and VoiceXML is a classic example in which the information
is held in {\tt <form>}s, and the dialog is expected to add data to the
{\tt <field>}s
in a form.  Each field can have a {\tt <prompt>} and a {\tt <grammar>} associated with
it. The VXML interpreter works its way through the form saying the prompt and
using the grammar to recognise what is said in response and putting it in the
field.
This is what happens when things go as expected, but a user may say
something that does not match a relevant grammar, or may say nothing at all.
To address this, VoiceXML provides two exception mechanisms, a
{\tt <nomatch>} element, the body of which is executed when the grammar doesn't
recognize what the user says, and a {\tt <noinput>} element for when the user
doesn't say anything.

This is the basic model but there are numerous other features, one of
particular interest is the variant that allows a matching grammar rule to fill
in more than one field.  For instance a prompt might ask "Where do you want to
fly to?" and the user might respond with "Brussels next Tuesday".  The system
providing this "mixed initiative" system would not only fill in the current
field with "Brussels" but also put Tuesday the 2nd of June in the appropriate
departure date field.  Note this is mixed initiative at the level of {\em data}
and, when one thinks of language as being only about information one may
question what else there is.  When we embrace the idea of intentions as first
order objects that can be talked about, the need for mixed initiative at
the level of intentions becomes apparent and TalkML was created to explicitly
provide this.

Voice XML was brilliantly conceived given that filling in forms is an
established interface between customers and a business process, and indeed
central to almost all internal business processes.  It was also theoretically
sound in that it is based on the notion of speech being about the information
it contained -- the dominant model at the time and probably still dominant
today.

\subsection{Talk Markup Language (TalkML)}
The TalkML interpreter works through a tkml file loading the static elements
and executing the action components in the order they appear.  Upon reaching a
{\tt <say recognize="...">} element, it loads the grammar associated with
the {\tt recognize} attribute and proceeds to say the body of the {\tt <say>} element while listening for a user utterance that matches the grammar.
After a certain period of time, possibly involving repair dialogs, the
interpreter moves on to the next say statement in the document until it
reaches the end of the document where upon the interpreter, in the case of
telephone based systems, hangs up.

This is the code for hello world:
\begin{verbatim}
<?xml version="1.0"?>
<tkml xmlns="http://www.cfpm.org/tkml">

  <achieves> say hello</achieves>

  Hello world!

</tkml>
\end{verbatim}

\noindent
Note that in this minimal hello world example the utterance to speak is not
wrapped in a {\tt<say/>} statement but, following the VoiceXML standard, the
textual content of a block is treated as if it should be said.  The
{\tt <achieves>} element is required and discussed below.

This TalkML code says hello and listens for the user saying hello back:
\begin{verbatim}
<?xml version="1.0"?>
<tkml xmlns="http://www.cfpm.org/tkml">

  <achieves>say hello</achieves>

  <grammar>
    <rule id="greeting">
      <one-of>
        <item>hello</item>
        <item>hi</item>
      </one-of>
    </rule>
  </grammar>

  <!-- the action starts here -->
  <say recognize="cid:local.greeting">
    Hello.
  </say>
  <say>
    Thank you for using this service.
    Goodbye.
  </say>
</tkml>
\end{verbatim}

\noindent

Normatively the user responds to what is said with something recognised by the
grammar rule which adds information to slots and the interpreter moves on to
the next action element.  This represents the seen-but-unnoticed type of
interaction in which the machine's conversational partner provides the second
part of an adjacency pair -- answers the question or says hello when greeted
and so on.  The key point of this paper is however not that hello-world is easy
to write -- there is certainly some ``verbage'' in this minimal code -- but
that the minimal code results in reasonable default behaviour when things do not go as hoped. 

\subsection{Default no-input behaviour}
In the doctor example above, the hearer - the patient - works hard at
accounting for the doctor's (non) utterance.  The success of the doctor's
communication act is not accidental; this is just how language works.
In order for the patient to ``work hard to account-for'' the doctor not
responding as expected -- not delivering the preferred second pair-part -- the
patent has to consider that the doctor and patient are actually {\em engaged}
in a conversation.  If the patient did not think that then there would be no
expectation of a response, there would be no need to account-for the (non)
response, and the doctor's message would not be communicated.  {\em Unlike} the
work done by calling someone's name across a crowded room, engagement is essential if a communicative act is going to be based on being noticed-and-accounted-for.

Setting up engagement is something that socialized humans know how to do
and that machines need to be instructed about~\cite{wallis10}.
Indeed modern speech recognition algorithms are based on the large
vocabulary continuous speech problem and rely on accurate ``end point''
detection which is quite difficult with a far field microphone where there is
likely to be background noise and indeed there may be background conversation.
Humans (and animals in general) do this using something like auditory scene
analysis which is technology still in the laboratory and a long way from being
commercially available. 
Engagement in a face to face conversation involves all sorts of cues from eye
contact to posture and these are indeed interesting but problematic if one's
interest is dialog.  Looking at dialog over the telephone mitigates all
these issues and indeed speech recognition has been optimized for the
telephone scenario given the massive commercial interest in phone based
service provision. 

Once a call is established the expectation is -- normative behaviour dictates
-- that the conversational partners will work through the process of
``closing'' the conversation~\cite{schsac73} before hanging up.  If the
conversation breaks down before that, for example if the caller doesn't say
anything, then such action needs accounting-for.  The human knows that.  So
if the machine assumes an explanation for no input, the human is {\em
compelled} to address the assumption.
Over the phone there is a very obvious explanation for why a conversational
response is not forthcoming and that is that the line is dead for some reason.
By having the system assume this, the human caller can account for its actions
and provide an explanation.  Consider:

  PROGRAM: The balance on that card is £10.43 (3 seconds silence)

  PROGRAM: Hello?

  HUMAN: one minute.

For the user to say nothing here would be to {\em not} cooperate in the
maintenance of the communicative process, and in a very real sense, the human
is {\em compelled} to correct the system's mistaken assumption.  If they are
still on the line, the human will say something.  Ideally the system would
understand the user's explanation for no response but it doesn't matter because
a reasonable system response to (almost) any user response is to simply wait a
while longer.  The user may say something that introduces a new issue but that
is handled by the {\tt no-match} mechanism below.  The point is that by having
the machine interpret the (non) act as one thing, the human conversational
partner {\em will} respond if they can.  They are compelled by the very nature
of being a socialized human being.  Rather than leaving it to the dialog
designer to think about what to do if nothing is said, TalKML (behaves as if the
system) has assumed the phone line is dead and tries to re-establish
engagement. If this fails it hangs up, but a human caller who did not respond
to the system's attempted repair will feel the system was justified in its
actions.

\subsection{Default no-match behaviour -- sanction}

The proposal is that the {\tt recognize} grammar associated with a {\tt <say} statement
matches the seen-but-unnoticed.  This is the normal behaviour or
``main game'' that the user is expected to follow.  Filling in a form for
instance might consist of a series of questions, each bracketed in a {\tt say}
statement referencing the {\tt grammar} that adds information to the appropriate
{\tt slot}.
The challenge is what to do if the human says something that is not
seen-but-unnoticed, but which the user (almost) certainly expects his or her
conversational partner to account for.  The premise is that accounting-for
requires something akin to intention recognition and this is discussed below,
but first consider how the system {\em ought}, as a social actor, to behave if
the accounting-for fails.  Under these conditions the human risks sanction and the machine needs to express that.

Threatening to sanction is of course highly dependent on cultural differences
and notions of power and distance.  Telling off a child is done differently to
telling off the football umpire and telling off someone else's child is
different in a different way.  Indeed in some cultures telling off is not done
by the mother but rather left to an aunty.
TalkML has been set up to, by default, start by apologising, and then move on to explaining why the human is being unreasonable.
From there the system threatens sanction by initiating a closing of the
conversation.  If the human remains unreasonable, the system initiates the
closing process and finally simply hangs up.  Consider this conversation in
which the TalkML script above gets a no-match response from the grammar:

  PROGRAM: Hello. Welcome to Peter's greeting service.

  HUMAN: have a banana.

  PROGRAM: I'm sorry, I didn't get that.

  HUMAN: banana?

  PROGRAM: The purpose of this system is to say hello.  How can I help?
  
  HUMAN: banana!
  
  PROGRAM: Thank you for using this service.
  
  HUMAN: Here, have a gorilla
  
  PROGRAM: Good bye.
  
  HUMAN: gorilla?

  PROGRAM: (hangs up)

\noindent
In order to do this conversation, the system only needs to be able to state, in
a broad sense, the services it offers -- that is it needs to be able to
summarise the goals it can achieve, and hence summarise what it is
reasonable to expect the system to do.  To this end the required element,
{\tt <achieves>}, has been added to the TalkKML element that completes the sentence
"The purpose of this system is to ..." so the machine can, at the top level, explain what is expected of the user.

This example might be extreme but it does demonstrate a critical point that we
all feel the machine is justified in sanctioning here; no one will think ''I
would not want to use the service on a regular basis''~\cite{CommEval} if he or
she had treated the machine in this way and got this response.  The challenge
is thus to characterize the purpose of the system in such a way as to delineate
what is reasonable -- to delineate behaviour the community of practice consider
is sanctionable from that which is not. We still need however to be able to
handle the utterances that members of the community of practice expect to be
accountable-for.

\subsection{Default no-match behaviour -- accounting-for}

Handling the noticed-and-accounted-for requires some form of recognition of
intent.  As discussed above, some cases where this is needed are trivial, and
others would appear impossible. The observation is that intention recognition
can be treated as an instance of the classic challenge of plan choice in a BDI
agent~\cite{clintPhD}. With a fixed set of plans, plan choice can be quite
``shallow'' and in the case of TalkML is based on what the user says.
It is a matter of empirical research to decide if the mechanism proposed is
enough; a research topic we leave for others with more resources.  

Where VoiceXML has {\tt form}s, TalkML has {\tt plan} elements. As with BDI
architectures, a plan is a static entity representing a sequence of actions
that achieve a goal.  This TalkML segment tells a knock-knock joke:
\begin{verbatim}
  <plan achieves="tell joke" >
    <say recognize="cid:local.whothere" > Knock knock.</say>
    <say recognize="cid:local.madamwho"> Madam </say>
    <say>Ma damn foot is stuck</say>
  </plan>
\end{verbatim}
To tell a knock-knock joke requires a sequence of five turns and is a good
example of a  case requiring a commitment to more than a single atomic
action.  As mentioned above plans are static and, following the BDI model, the
system forms an {\em intention} only when the goal is {\tt post}ed and the
system finds a plan that might achieve that goal.  According to the folk
psychological model upon which BDI is based~\cite{BIP} humans do similar
thing, and the aim of an intention recognition mechanism is to identify
the goal/plan the human is using.  The observation is that there is no point
recognizing the plans and goals of the human if the system does not have a
plan for dealing with the situation.  The challenge for the dialog
designer is to come up with a set of plans that deals with all the cases
for which sanction by the system would not be, in the eyes of members of
the community of practice, justified.  The aim here is to provide the tools
that allow the dialog designer to systematically address that space.

The proposal is that the intent of the user can be recognised from what the
user says, and recognising what the user says is done via a {\tt trigger}
grammar.  In theory the trigger grammar would be associated with a plan or goal
in a BDI architecture representing the user, and the system, also using a BDI
architecture, would choose which goal/plan to use based on reasoning about the
state of the model of the user.  However this seems overly complicated given
the type of dialog one might expect to be able to deal with and so trigger
grammars are associated with the system plans rather than with user plans and
goals.  There is a case for having trigger grammars associated with system
goals but again, as a first pass, TalkML provides triggers only for system
plans.  Figure~\ref{kk} is the full code (with external grammars) for a
knock-knock joke teller that provides {\em mixed initiative at the level of
intentions}.
\begin{figure*}
\begin{verbatim}
<?xml version="1.0"?>
<tkml version="0.1" xmlns="http://www.cfpm.org/tkml">
  <achieves>tell a joke</achieves>

  <plan achieves="tell joke">
    <say recognize="cid:kk.whosthere">Knock knock.</say>
    <say recognize="cid:kk.shoewho">Wooden shoe</say>
    <say>Wooden shoe like to hear another joke?</say>
  </plan>

  <plan achieves="say goodbye" trigger="cid:util.ouc">
    <say recognize="cid:util.bye">Thanks for using this service.</say>
    <say> Good bye. </say>
    <hangup/>
  </plan>

  <grammar id="util" src="utilities.srgs"/>
  <grammar id="kk" src="knockknock.srgs"/>

<!-- action starts here -->
<say recognize="cid:util.yesNo">
  Hello. Want to hear a joke?  </say>
<if cond="$userSaid==yes">
  <post goal="tell joke"/>
<else/>
  <say> Oh, Okay. </say>
</if>
<post goal="say goodbye"/>
</tkml>

\end{verbatim}

\caption{A knock-knock joke teller with 2 plans.} \label{kk}
\end{figure*}

Here are two ways to {\em not} hear the joke.
\begin{verbatim}
  System: Hello. Want to hear a joke?
  User:   No.
  System: Thanks for using this service.
  User:   Bye
  System: Good bye.
\end{verbatim}

\noindent
In this case control follows the logic of the program, which gives the user the
option of saying no. In this case
\begin{verbatim}
  System: Hello. Want to hear a joke?
  User:   Bye
  System: Good bye.
\end{verbatim}

\noindent
The user initiates a plan that is not in the sequence specified by the program
but rather says something that needs accounting-for. It works by loading a
second grammar consisting of the trigger attributes of plans, associated with
the plan to which the trigger is attached. If the grammar specified in a {\tt
recognize} attribute of  {\tt say} element results in a no-match, the system
tries the user's utterance against this {\em trigger grammar}.  If there is a
match, then the assumption is that the user is initiating a new plan for which
the system's referenced plan is appropriate.  In effect the system's response
to a no-match is to accounts-for the user's (not seen-but-unnoticed) utterance by assuming a change of intent.

The code could have been written the other way with a trigger for telling the
joke being a grammar that matches, among other things, ``do you know any
jokes?''. The action part of the script would not have to reference the joke
plan - the body of the script can simply be ``the main game'' dealing with bank
transactions or something.  Providing users with the ability to introduce their
own goals {\em without} changing the logic of the dialog script not only allows
for mixed initiative at the level of intentions, it also results in extremely
modular code.

\section{Conclusion}
In this paper we advocate a holistic approach in which, rather than looking in
heads for meaning, we take seriously the notion of talk as action~\cite{austin,searle69} in a social space rather than as a conduit for information~\cite{Reddy93}.
It seems talk is full of protocols that language based systems must use if they are to be considered as other than a (rather awkward) computer interface~\cite{ibtbagm}.

The paper introduced the idea from the CA community of language as
interaction in a community of practice, and the notion that a speaker's
utterance will go {\em seen but unnoticed}, {\em noticed and accounted for}, or
{\em risk sanction}.  We know how to do the first, but our conversational
machines regularly end up doing the last.  In order to implement a process of
accounting for, the paper introduces a limited means of intention recognition
based on a BDI agent architecture running a fixed plan library.
Talk of intentions may (or may not~\cite{Denn87}) be unscientific, but it is
how everyday folk think of other people and is built in to the workings of
human languages.  The resulting implementation provides what can be thought of
as {\em mixed initiative at the level of intentions} in that the human
conversational partner can introduce new goals at any time, even if the machine
is part way through executing its current plan.

Taking seriously the idea that the machine is a social actor we conclude that
the machine is not only allowed to sanction but expected to sanction. This
provides a means of keeping the user on topic; it is up to the dialog designer
to decide how and when to threaten sanction.

In combination these two result in clean code where the executable section of a
tkml script or plan is simply the ``main game'' with progress through the
script halted while repair takes place, or the (hidden) goal trigger mechanism
jumping execution to appropriate plan.

\section{Acknowledgements}
 All images are in the public domain without restrictions, available through
Wikimedia, with the exception of the sailing yacht in Figure 3 which is
licensed under the Creative Commons Attribution-Share Alike to Anco Kok.
The bar-plot in Figure 7 first appeared in \cite{wcl14}.

%Images in Section~\ref{daass} are from the Creative Commons with the following
%URLs:
% https://commons.wikimedia.org/wiki/File:Francesco_vinea_painting1.jpg
% https://commons.wikimedia.org/wiki/File:Sailing_Yacht_Axonite,_designed_by_Guido_de_Groot.jpg
%https://commons.wikimedia.org/wiki/File:Steam_locomotive_%22Bodiam%22.jpg
%https://commons.wikimedia.org/wiki/File:Brain_headBorder.jpg
%or self published in the public domain:
% sectioned_loco2.png
% barplot.png (appeared first in Peter Wallis and Keeley Crockett and Clare
% Little, "When Things Go Wrong", in Human-Agent Interaction Design and Models (HAIDM), Sarvapali D. Ramchurn and Joel Fisher and Avi Rosenfeld and Long Tran-Thanh and Kobi Gal, Paris, AAMAS 2012
% The algorithm in Figure~\ref{acl} is from Wooldridge's book

%\bibliographystyle{abbrv}
%\bibliography{../../mybib.bib} 

\end{document}